\documentclass{article}

\usepackage{graphicx}
\graphicspath{./figures/}
\usepackage[export]{adjustbox}

\usepackage{amsmath}
\usepackage{amssymb}
\usepackage{booktabs}

\usepackage{multirow}
\usepackage{graphicx}  
\usepackage{bm}        
\usepackage{booktabs}
\usepackage[normalem]{ulem}  

\title{InvDec: Inverted Decoder for Multivariate Time Series Forecasting 
with Separated Temporal and Variate Modeling}

\author{
    Yuhang Wang \\
    \textit{Department of Computer Science} \\
    \textit{Hangzhou City University} \\
    \texttt{2240101022@stu.hzcu.edu.cn}
}

\date{\today}

\begin{document}

	\maketitle

	\begin{abstract}
        Multivariate time series forecasting requires simultaneously modeling temporal patterns and cross-variate dependencies. Channel-independent methods such as PatchTST excel at temporal modeling but ignore variable correlations, while pure variate-attention approaches such as iTransformer sacrifice temporal encoding. We propose \textbf{InvDec} (Inverted Decoder), a hybrid architecture that achieves principled separation between temporal encoding and variate-level decoding. InvDec combines a patch-based temporal encoder with an inverted decoder operating on the variate dimension through variate-wise self-attention. We introduce delayed variate embeddings that enrich variable-specific representations only after temporal encoding, preserving temporal feature integrity. An adaptive residual fusion mechanism dynamically balances temporal and variate information across datasets of varying dimensions. Instantiating InvDec with PatchTST yields InvDec-PatchTST. Extensive experiments on seven benchmarks demonstrate significant gains on high-dimensional datasets: 20.9\% MSE reduction on Electricity (321 variables), 4.3\% improvement on Weather, and 2.7\% gain on Traffic compared to PatchTST, while maintaining competitive performance on low-dimensional ETT datasets. Ablation studies validate each component, and analysis reveals that InvDec's advantage grows with dataset dimensionality, confirming that cross-variate modeling becomes critical as the number of variables increases.
    \end{abstract}


	\newpage
    
    \section{Introduction}
    
    Time series forecasting is a fundamental problem with widespread applications across diverse domains, including energy management, transportation planning, climate science, healthcare monitoring, and financial market analysis~\cite{kim2025comprehensivesurveydeeplearning,Kong_2025}. The ability to accurately predict future trends and patterns from historical observations is critical for decision-making, resource allocation, and risk management in these fields~\cite{qiu2025comprehensivesurveydeeplearning}. Traditional statistical approaches such as ARIMA~\cite{Box2015} have long served as the foundation for time series analysis, offering interpretable models based on stationarity assumptions and linear relationships. However, as real-world time series data have grown increasingly complex—characterized by high dimensionality, non-stationarity, and intricate temporal dependencies—the limitations of classical methods have become apparent~\cite{kim2025comprehensivesurveydeeplearning}.
    
    The advent of deep learning has revolutionized time series forecasting, enabling models to capture non-linear patterns and long-range dependencies that elude traditional techniques~\cite{Kong_2025}. Among deep learning architectures, Transformers~\cite{vaswani2023attentionneed} have emerged as particularly powerful, leveraging self-attention mechanisms to model temporal dynamics across extended horizons. Early Transformer-based forecasting models such as Informer~\cite{zhou2021informerefficienttransformerlong}, Autoformer~\cite{wu2022autoformerdecompositiontransformersautocorrelation}, and FEDformer~\cite{zhou2022fedformerfrequencyenhanceddecomposed} focused primarily on improving computational efficiency and capturing temporal patterns through innovations such as sparse attention and decomposition strategies. Despite their success, these models largely adopt a \textbf{channel-independent} paradigm, treating each variable in a multivariate time series as an isolated entity~\cite{qiu2025comprehensivesurveydeeplearning}. This design choice, while computationally efficient, overlooks the rich correlations and interactions among variables—a critical limitation in high-dimensional forecasting scenarios where cross-variate dependencies often encode essential information about underlying system dynamics~\cite{qiu2025comprehensivesurveydeeplearning}.
    
    Recent research has begun to address this gap by exploring architectures that explicitly model relationships across the variate dimension. Notably, methods such as iTransformer~\cite{liu2024itransformerinvertedtransformerseffective} invert the traditional Transformer design by applying attention mechanisms on variates rather than time steps, while others such as Crossformer~\cite{zhang2023crossformer} interleave temporal and variate modeling throughout the network. Although these approaches demonstrate the importance of cross-variate dependencies, they often involve trade-offs: some sacrifice fine-grained temporal modeling, while others introduce architectural complexity that complicates interpretation and increases computational overhead. Furthermore, specialized methods designed for exogenous forecasting~\cite{wang2024timexer} or spatial grid data~\cite{cheng2024leveraging2dinformationlongterm} address domain-specific challenges but do not generalize to arbitrary multivariate time series where variables lack explicit spatial topology or external covariates.
    
    In this paper, we propose \textbf{InvDec} (\textbf{Inv}erted \textbf{Dec}oder), a novel hybrid architecture that achieves a principled separation between temporal encoding and variate decoding. InvDec builds upon the strengths of patch-based temporal modeling~\cite{nie2023timeseriesworth64} while introducing an inverted decoder that operates on the variate dimension to capture cross-variate dependencies. Unlike prior methods that entangle temporal and variate modeling or sacrifice one for the other, InvDec explicitly decouples these objectives: the encoder focuses solely on extracting temporal features from each variable's sequence, while the decoder performs variate-level attention to model inter-variable correlations. To further enrich variate representations, we incorporate learnable variate embeddings~\cite{wang2024timexer} in the decoder, but crucially delay their introduction until after temporal encoding to avoid distorting temporal patterns. Additionally, an adaptive residual fusion mechanism dynamically balances contributions from temporal and variate pathways, enabling the model to adapt to datasets of varying dimensions and correlation structures.

    \begin{figure}[t]
        \centering
        \includegraphics[width=0.5\textwidth]{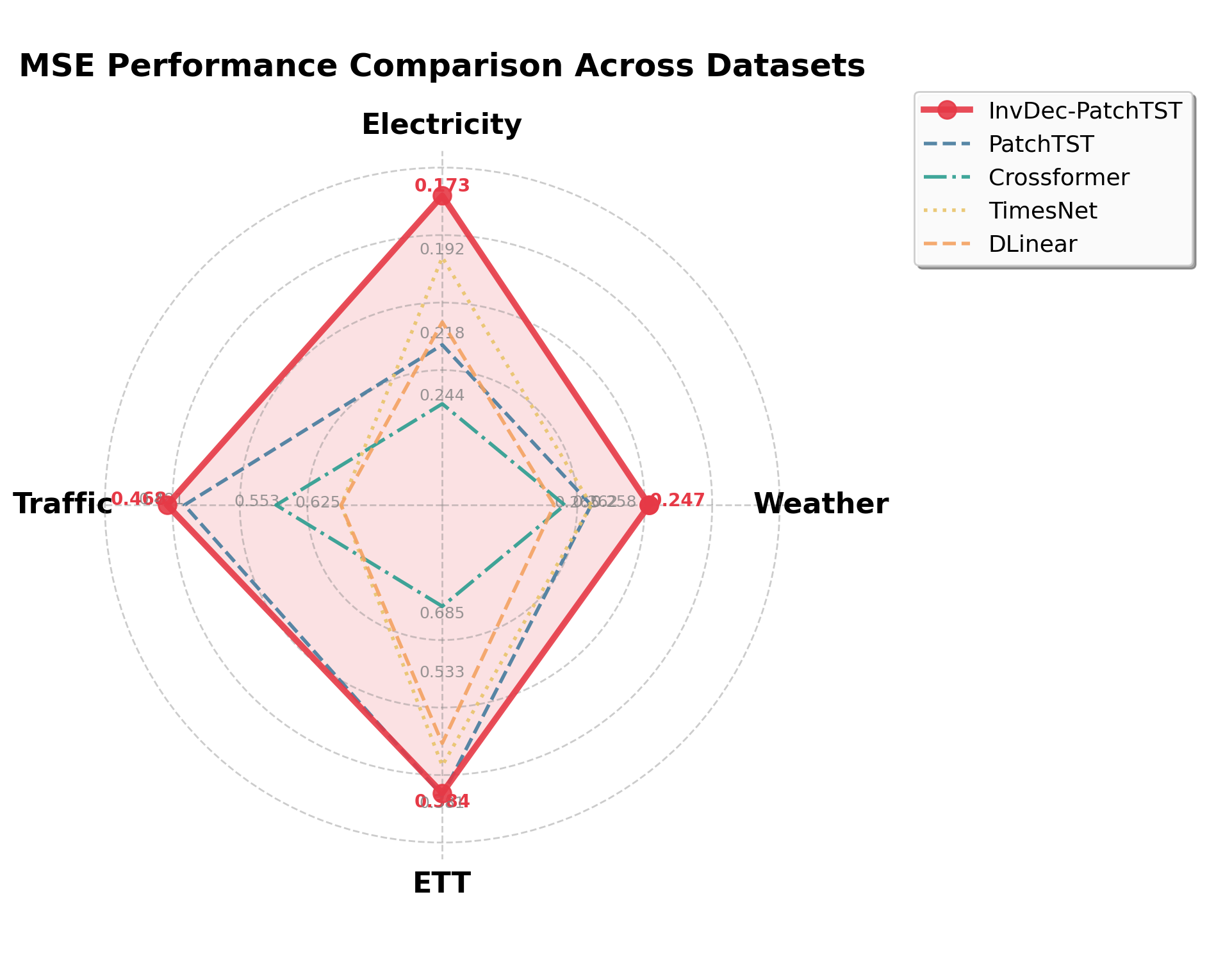}
        \caption{Performance overview: InvDec-PatchTST (red) consistently outperforms baselines on high-dimensional datasets (Electricity, Weather, Traffic) while remaining competitive on low-dimensional ETT benchmarks. Radial distance represents normalized MSE.}
        \label{fig:radar_intro}
    \end{figure}
    
    Our contributions in this work are threefold. First, we introduce InvDec, a hybrid encoder-inverted decoder architecture that, to the best of our knowledge, represents the first work to combine a temporal encoder with a variate-level decoder in this manner, achieving superior performance on multivariate forecasting tasks through clear separation of modeling concerns. Second, we propose a delayed variate embedding strategy that enriches variate-specific representations without interfering with temporal encoding, alongside an adaptive residual fusion mechanism that enables dynamic balancing of temporal and variate information across diverse datasets. Third, we conduct extensive experiments on seven widely-used benchmarks, demonstrating that InvDec achieves significant performance gains on high-dimensional datasets—including a 20.9\% MSE reduction on Electricity (321 variables) compared to PatchTST—while maintaining competitive performance on low-dimensional datasets (Figure~\ref{fig:radar_intro}), with comprehensive ablation studies validating the effectiveness of each architectural component.
  
    \section{Related Work}
    \label{sec:related}
    
    \subsection{Deep Learning Architectures for Time Series Forecasting}
    
    Prior to the widespread adoption of Transformers, deep learning approaches to time series forecasting relied primarily on recurrent neural networks (RNNs), convolutional neural networks (CNNs), and multi-layer perceptrons (MLPs)~\cite{kim2025comprehensivesurveydeeplearning,Kong_2025}. RNN-based architectures, including LSTMs and GRUs, were designed to capture sequential dependencies through recurrent connections, enabling them to model temporal dynamics in univariate and multivariate settings~\cite{DBLP:journals/eswa/HuangLZLLZ24}. However, these methods suffer from training instability, vanishing gradients, and limited capacity to capture long-range dependencies~\cite{kim2025comprehensivesurveydeeplearning}. CNN-based approaches such as Temporal Convolutional Networks (TCN)~\cite{bai2018empiricalevaluationgenericconvolutional} leverage dilated convolutions to expand receptive fields, achieving competitive performance while maintaining computational efficiency. Recent works such as TimesNet~\cite{wu2023timesnettemporal2dvariationmodeling} and ModernTCN~\cite{donghao2024moderntcn} further demonstrate the effectiveness of convolutional architectures by exploiting multi-scale temporal patterns and hierarchical feature extraction.
    
    Graph Neural Networks (GNNs) have also been explored for time series forecasting, particularly in scenarios where variables exhibit explicit spatial or relational structures~\cite{jin2024surveygraphneuralnetworks}. Methods such as spatial-temporal graph convolutional networks~\cite{yan2018spatialtemporalgraphconvolutional} and temporal graph networks~\cite{rossi2020temporalgraphnetworksdeep} model interactions among variables through graph attention mechanisms~\cite{veličković2018graph}. While effective for spatially-structured data (e.g., traffic networks, sensor grids), these approaches require predefined graph topologies and may not generalize to arbitrary multivariate time series where relationships among variables are latent and dynamic.
    
    In parallel, MLP-based architectures have re-emerged as surprisingly competitive alternatives to more complex models. DLinear~\cite{zeng2022transformerseffectivetimeseries} challenges the necessity of Transformers by demonstrating that simple linear mappings can outperform sophisticated architectures on certain benchmarks. Building on this insight, methods such as N-BEATS~\cite{Oreshkin2020N-BEATS:}, N-HiTS~\cite{challu2022nhitsneuralhierarchicalinterpolation}, TSMixer~\cite{Ekambaram_2023}, and TimeMixer~\cite{wang2024timemixerdecomposablemultiscalemixing} achieve strong performance through hierarchical decomposition, multi-scale mixing, and efficient parameter utilization. These works highlight the importance of inductive biases and architectural simplicity, though they often lack explicit mechanisms for modeling cross-variate dependencies in high-dimensional settings~\cite{qiu2025comprehensivesurveydeeplearning}.
    
    \subsection{Transformer-based Time Series Forecasting}
    
    The introduction of Transformers~\cite{vaswani2023attentionneed} revolutionized sequence modeling through self-attention mechanisms that enable parallelized training and flexible modeling of long-range dependencies. Early adaptations to time series forecasting, such as Informer~\cite{zhou2021informerefficienttransformerlong}, addressed the quadratic complexity of self-attention by proposing sparse attention mechanisms, enabling efficient processing of long sequences. Subsequent works further refined Transformer-based architectures through innovations such as auto-correlation mechanisms in Autoformer~\cite{wu2022autoformerdecompositiontransformersautocorrelation}, frequency-domain enhancements in FEDformer~\cite{zhou2022fedformerfrequencyenhanceddecomposed}, and stationarity-aware normalization in Non-stationary Transformers~\cite{liu2023nonstationarytransformersexploringstationarity}. These methods primarily focus on temporal modeling, treating each variable independently to reduce computational overhead—a strategy often referred to as the \textbf{channel-independent} paradigm~\cite{qiu2025comprehensivesurveydeeplearning}.
    
    A critical advancement in this direction came with PatchTST~\cite{nie2023timeseriesworth64}, which introduced patch-based tokenization inspired by Vision Transformers. By segmenting time series into non-overlapping patches and treating each patch as a token, PatchTST significantly reduces sequence length and computational cost while achieving state-of-the-art performance on numerous benchmarks. However, its strict channel-independence assumption—where each variable is processed in isolation—limits its ability to capture cross-variate correlations, particularly in high-dimensional scenarios where such dependencies are critical for accurate forecasting~\cite{qiu2025comprehensivesurveydeeplearning}. Extensions such as PETformer~\cite{lin2023petformerlongtermtimeseries} and multi-resolution Transformers~\cite{zhang2024multiresolutiontimeseriestransformerlongterm} build upon patching strategies with enhancements for placeholder tokens and hierarchical temporal resolutions, but largely maintain the channel-independent paradigm.
    
    Complementary innovations have explored multi-scale modeling and frequency-domain enhancements. Scaleformer~\cite{shabani2023scaleformer} and Pathformer~\cite{chen2024pathformer} introduce iterative refinement and adaptive pathways to capture temporal patterns at multiple resolutions, while Fredformer~\cite{Piao_2024} and BasisFormer~\cite{ni2024basisformerattentionbasedtimeseries} leverage frequency decomposition and learnable basis functions to enhance feature representations. These methods demonstrate the value of multi-scale and frequency-aware designs, yet do not fundamentally address the challenge of modeling cross-variate dependencies.
    
    \subsection{Cross-Variate Modeling and Novel Forecasting Paradigms}
    
    Recognizing the limitations of channel-independent approaches, recent research has explored architectures that explicitly model relationships across the variate dimension~\cite{qiu2025comprehensivesurveydeeplearning}. Crossformer~\cite{zhang2023crossformer} proposes a two-dimensional attention mechanism that interleaves temporal and variate modeling through cross-dimension dependency blocks. While this design captures correlations among variables, the entanglement of temporal and variate attention complicates interpretation and increases computational complexity. DSformer~\cite{yu2023dsformerdoublesamplingtransformer} adopts a double-sampling strategy to balance temporal and spatial dependencies, but similarly lacks a clear separation between these modeling objectives.
    
    A paradigm shift emerged with iTransformer~\cite{liu2024itransformerinvertedtransformerseffective}, which inverts the traditional Transformer design by applying self-attention on the variate dimension rather than the temporal dimension. By treating each variable as a token, iTransformer excels at capturing cross-variate dependencies, particularly in ultra-high-dimensional datasets (e.g., Traffic with 862 variables). However, this design completely sacrifices fine-grained temporal modeling, as attention is no longer applied across time steps. Other methods explore variate-level correlations through specialized mechanisms: VCformer~\cite{yang2024vcformervariablecorrelationtransformer} models lagged correlations among variables, UniTST~\cite{liu2024unitsteffectivelymodelinginterseries} balances inter-series and intra-series dependencies, and DeformTime~\cite{shu2025deformtimecapturingvariabledependencies} employs deformable attention to adaptively capture variable interactions. While these approaches demonstrate the importance of cross-variate modeling, they often involve trade-offs between temporal and variate perspectives or introduce architectural complexity that hinders scalability and interpretation.
    
    Specialized forecasting paradigms have also been proposed for domain-specific challenges. TimeXer~\cite{wang2024timexer} introduces variate-specific embeddings to handle heterogeneity in exogenous forecasting tasks, where external covariates (e.g., weather, holidays, prices) are used to predict target variables. By fusing variable embeddings at the input stage, TimeXer addresses the systematic time lags and diverse measurement units inherent in exogenous variables. However, its focus on exogenous forecasting limits its applicability to standard multivariate forecasting benchmarks, where all variables are endogenous and correlations are dynamic rather than statically heterogeneous. Similarly, methods designed for spatial grid data~\cite{cheng2024leveraging2dinformationlongterm} leverage two-dimensional information to model spatially-structured variables (e.g., traffic networks, sensor grids), but do not generalize to arbitrary multivariate time series where variables lack explicit spatial topology.
    
    In contrast to these approaches, our proposed InvDec architecture achieves a principled separation between temporal encoding and variate decoding. Unlike iTransformer, which abandons temporal attention, InvDec retains a patch-based temporal encoder to capture fine-grained temporal patterns. Unlike Crossformer, which interleaves temporal and variate modeling, InvDec explicitly decouples these objectives through a hybrid encoder-decoder design. Unlike TimeXer, which fuses variable embeddings at the input stage, InvDec delays the introduction of variate embeddings until after temporal encoding, preserving the integrity of temporal features while enriching variate-specific representations. This design enables InvDec to leverage the strengths of both temporal and variate modeling without sacrificing either, achieving superior performance on high-dimensional datasets while maintaining competitive results on low-dimensional benchmarks.
    
    \section{Methodology}
        
    \subsection{Problem Formulation}
    \label{subsec:problem}
    
    Let $\mathbf{X} \in \mathbb{R}^{L \times C}$ denote a multivariate time series consisting of $L$ time steps and $C$ variables. The input can be represented as a concatenation of $C$ individual variable sequences:
    \begin{equation}
    \mathbf{X} = [\mathbf{x}_1, \mathbf{x}_2, \dots, \mathbf{x}_C],
    \end{equation}
    where $\mathbf{x}_c \in \mathbb{R}^L$ is the complete temporal sequence of the $c$-th variable for $c \in \{1, \dots, C\}$.
    
    The objective of long-term forecasting is to predict the values for the next $H$ time steps based on the $L$ historical observations:
    \begin{equation}
    \hat{\mathbf{Y}} = \mathcal{F}(\mathbf{X}; \Theta),
    \end{equation}
    where $\mathcal{F}(\cdot)$ represents the forecasting model parameterized by $\Theta$. The model is trained by minimizing the Mean Squared Error (MSE) or Mean Absolute Error (MAE) between the predicted values $\hat{\mathbf{Y}} = [\hat{\mathbf{y}}_1, \hat{\mathbf{y}}_2, \dots, \hat{\mathbf{y}}_C] \in \mathbb{R}^{H \times C}$ and the ground truth future values $\mathbf{Y} = [\mathbf{y}_1, \mathbf{y}_2, \dots, \mathbf{y}_C] \in \mathbb{R}^{H \times C}$.
        
    \subsection{InvDec: Inverted Decoder Architecture}
    \label{subsec:invdec}

    \begin{figure}[ht]
        \centering
        \includegraphics[width=0.65\textwidth]{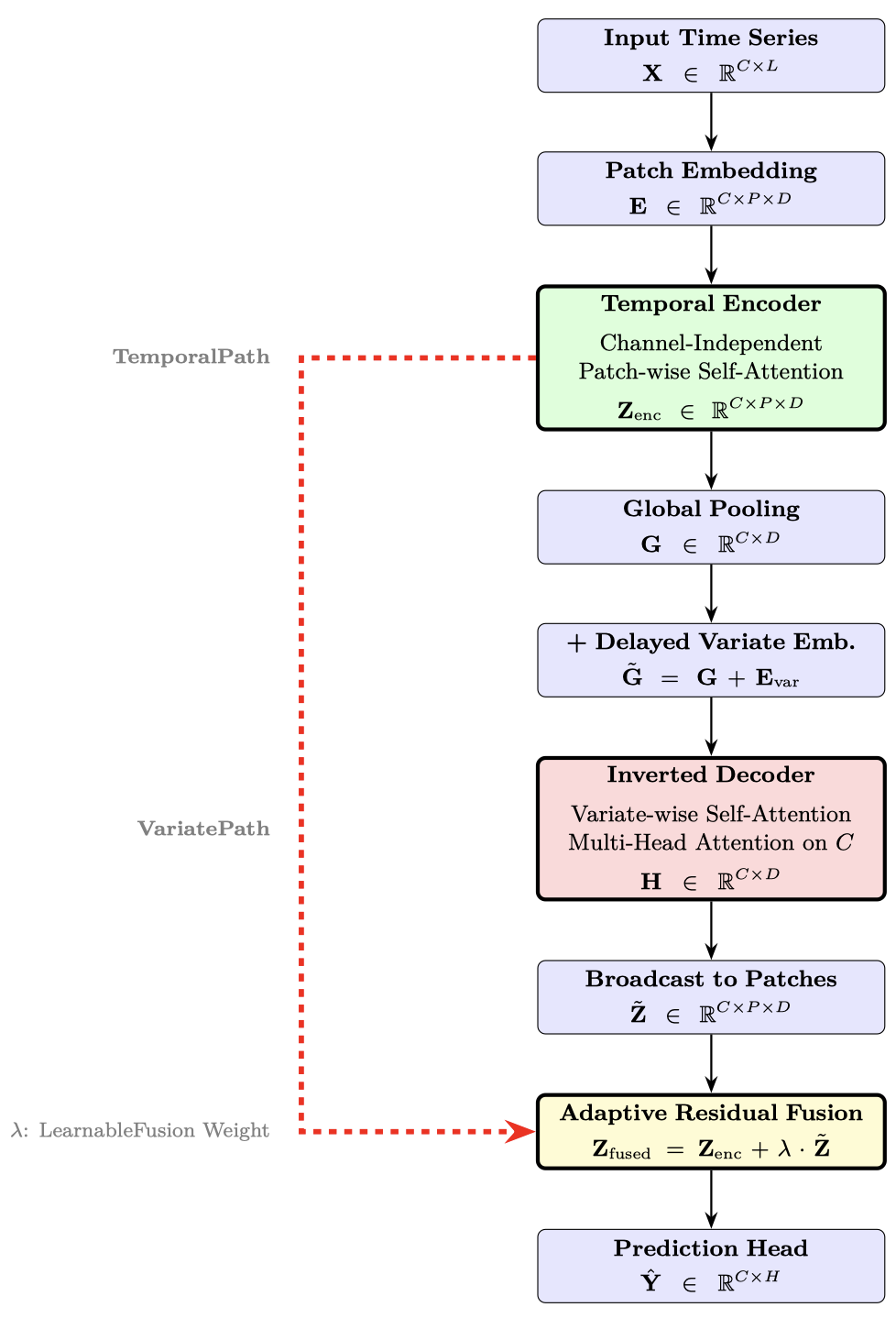}
        \caption{Architecture of InvDec-PatchTST. The temporal encoder (green) captures patch-wise temporal patterns in a channel-independent manner, while the inverted decoder (red) models cross-variate dependencies through variate-wise self-attention. Delayed variate embeddings $\mathbf{E}_{\textrm{var}}$ are introduced only after temporal encoding to preserve temporal feature integrity. An adaptive residual fusion mechanism (yellow, red dashed path) with learnable weight $\lambda$ dynamically balances contributions from both pathways.}
        \label{fig:architecture}
    \end{figure}
    
    We propose \textbf{InvDec} (\textbf{Inv}erted \textbf{Dec}oder), a hybrid architecture that combines temporal encoding and variate-level decoding to simultaneously capture temporal patterns and cross-variate dependencies. Unlike pure inverted architectures such as iTransformer~\cite{liu2024itransformerinvertedtransformerseffective}, which apply attention mechanisms solely on the variate dimension, and unlike cross-dimension methods such as Crossformer~\cite{zhang2023crossformer}, which interleave temporal and variate perspectives throughout the model, InvDec maintains a \textbf{clear separation of concerns}: the encoder focuses on temporal modeling, while the decoder focuses on dependencies across the variate dimension.
    
    To validate the effectiveness of InvDec, we instantiate it using PatchTST~\cite{nie2023timeseriesworth64} as the temporal modeling backbone, yielding the \textbf{InvDec-PatchTST} architecture, as is illustrated in Figure~\ref{fig:architecture}. PatchTST has demonstrated great performance on multiple forecasting benchmarks through its efficient patch embedding and channel-independent modeling strategy. However, its channel-independence assumption ignores correlations among variables, limiting its performance in high-dimensional multivariate scenarios. By introducing the InvDec decoder after PatchTST's encoder, InvDec-PatchTST retains the strong temporal modeling capability of PatchTST while explicitly capturing cross-variate dependencies. This design enables the model to simultaneously leverage local temporal patterns (via the patching mechanism) and global variate correlations (via variate-level attention), achieving superior performance on complex multivariate forecasting tasks. It is worth noting that InvDec is a plug-and-play module that could theoretically be combined with other temporal modeling methods (e.g., TimesNet~\cite{wu2023timesnettemporal2dvariationmodeling}, Informer~\cite{zhou2021informerefficienttransformerlong}), though we choose PatchTST as a representative implementation for systematic evaluation.
    
    \textbf{Patch Embedding Layer.} Following PatchTST's design, we first apply a patch-based segmentation strategy to the input time series $\mathbf{X}$. For the $c$-th variable where $c \in \{1, \dots, C\}$, the univariate sequence $\mathbf{x}_c \in \mathbb{R}^L$ is divided into $P$ non-overlapping patches:
    \begin{equation}
    \mathbf{x}_c = [\mathbf{p}_{c,1}, \mathbf{p}_{c,2}, \dots, \mathbf{p}_{c,P}],
    \end{equation}
    where $\mathbf{p}_{c,i} \in \mathbb{R}^{S}$ for $i \in \{1, \dots, P\}$, $S$ is the patch length, and $P = \lfloor \frac{L - S}{s} \rfloor + 1$ with stride $s$. Each patch is then linearly projected into a $D$-dimensional embedding space:
    \begin{equation}
    \mathbf{e}_{c,i} = \mathbf{W}_p \mathbf{p}_{c,i} + \mathbf{b}_p \in \mathbb{R}^D,
    \end{equation}
    where $\mathbf{W}_p \in \mathbb{R}^{D \times S}$ and $\mathbf{b}_p \in \mathbb{R}^D$ are learnable parameters. Finally, we stack the patch embeddings across all variables to obtain the patch embedding tensor:
    \begin{equation}
    \mathbf{E} = [\mathbf{e}_{1,:}; \mathbf{e}_{2,:}; \dots; \mathbf{e}_{C,:}] \in \mathbb{R}^{C \times P \times D},
    \end{equation}
    where $\mathbf{e}_{c,:} = [\mathbf{e}_{c,1}, \mathbf{e}_{c,2}, \dots, \mathbf{e}_{c,P}] \in \mathbb{R}^{P \times D}$ denotes all patch embeddings for the $c$-th variable.
    
    \textbf{Temporal Encoder.} The encoder independently processes each variable to capture temporal dependencies. Before applying $L_{\text{enc}}$ layers of self-attention, we reshape $\mathbf{E}$ into a $(C \cdot P) \times D$ matrix:
    \begin{equation}
    \mathbf{Z}^{(0)} = \text{Reshape}(\mathbf{E}, [C \cdot P, D]).
    \end{equation}
    For each encoder layer $\ell \in \{1, \dots, L_{\text{enc}}\}$, we compute:
    \begin{align}
    \mathbf{Q}^{(\ell)}, \mathbf{K}^{(\ell)}, \mathbf{V}^{(\ell)} &= \mathbf{Z}^{(\ell-1)} \mathbf{W}_Q^{(\ell)}, \mathbf{Z}^{(\ell-1)} \mathbf{W}_K^{(\ell)}, \mathbf{Z}^{(\ell-1)} \mathbf{W}_V^{(\ell)}, \\
    \mathbf{A}^{(\ell)} &= \text{softmax}\left(\frac{\mathbf{Q}^{(\ell)} (\mathbf{K}^{(\ell)})^\top}{\sqrt{d_k}}\right) \mathbf{V}^{(\ell)}, \\
    \mathbf{Z}^{(\ell)} &= \text{LayerNorm}(\mathbf{Z}^{(\ell-1)} + \text{Dropout}(\mathbf{A}^{(\ell)})), \\
    \mathbf{Z}^{(\ell)} &= \text{LayerNorm}(\mathbf{Z}^{(\ell)} + \text{FFN}(\mathbf{Z}^{(\ell)})),
    \end{align}
    where $\mathbf{W}_Q^{(\ell)}, \mathbf{W}_K^{(\ell)}, \mathbf{W}_V^{(\ell)} \in \mathbb{R}^{D \times D}$ are projection weight matrices, $d_k = \frac{D}{N_h}$ is the dimension per attention head with $N_h$ heads, and all intermediate variables $\mathbf{Q}^{(\ell)}, \mathbf{K}^{(\ell)}, \mathbf{V}^{(\ell)}, \mathbf{A}^{(\ell)}, \mathbf{Z}^{(\ell)} \in \mathbb{R}^{(C \cdot P) \times D}$. The final encoder output is obtained by reshaping:
    \begin{equation}
    \mathbf{Z}_{\text{enc}} = \text{Reshape}(\mathbf{Z}^{(L_{\text{enc}})}, [C, P, D]),
    \end{equation}
    where $(\mathbf{Z}_{\text{enc}})_{c,i,:}$ represents the feature representation of the $i$-th patch of the $c$-th variable after temporal modeling.
    
    \textbf{Inverted Decoder.} The core of our architecture, the inverted decoder operates on the \textbf{variate dimension} to capture cross-variate dependencies. This design is motivated by the observation that variables in high-dimensional multivariate time series often exhibit complex correlations (e.g., traffic flows across different locations, electricity consumption across regions). First, we extract the patch representations $\mathbf{z}_{c,i} = (\mathbf{Z}_{\text{enc}})_{c,i,:}$ from the encoder output and aggregate temporal information for each variable via global pooling across patches:
    \begin{equation}
    \mathbf{g}_c = \frac{1}{P} \sum_{i=1}^{P} \mathbf{z}_{c,i} \in \mathbb{R}^D, \quad c \in \{1, \dots, C\},
    \end{equation}
    yielding $\mathbf{G} \in \mathbb{R}^{C \times D}$. To encode the unique semantic information and physical characteristics of each variable (e.g., physical properties of different sensors, climatic features of different geographic locations), inspired by other related embedding methods~\cite{wang2024timexer}~\cite{zelaszczyk2024interpretablemultitasklearningshared}, we introduce \textbf{learnable variable embeddings} $\mathbf{E}_{\textrm{var}} \in \mathbb{R}^{C \times D}$ as variate-level representation enhancement, analogous to positional encodings in Transformers but operating on the variate dimension:
    \begin{equation}
    \tilde{\mathbf{G}} = \mathbf{G} + \mathbf{E}_{\text{var}},
    \end{equation}
    where $\mathbf{E}_{\text{var}}$ is initialized with small random values (e.g., $\mathcal{N}(0, 0.02^2)$) and optimized during training. This enables the model to learn a unique representation for each variable, capturing structured relationships among variables. Next, we apply $L_{\text{dec}}$ layers of variate-wise self-attention. Let $\mathbf{H}^{(0)} = \tilde{\mathbf{G}}$. For each decoder layer $\ell \in \{1, \dots, L_{\text{dec}}\}$:
    \begin{align}
    \mathbf{Q}_v^{(\ell)}, \mathbf{K}_v^{(\ell)}, \mathbf{V}_v^{(\ell)} &= \mathbf{H}^{(\ell-1)} \mathbf{W}_{Q_v}^{(\ell)}, \mathbf{H}^{(\ell-1)} \mathbf{W}_{K_v}^{(\ell)}, \mathbf{H}^{(\ell-1)} \mathbf{W}_{V_v}^{(\ell)}, \\
    \mathbf{A}_v^{(\ell)} &= \text{softmax}\left(\frac{\mathbf{Q}_v^{(\ell)} (\mathbf{K}_v^{(\ell)})^\top}{\sqrt{d_k}}\right) \mathbf{V}_v^{(\ell)}, \\
    \mathbf{H}^{(\ell)} &= \text{LayerNorm}(\mathbf{H}^{(\ell-1)} + \text{Dropout}(\mathbf{A}_v^{(\ell)})), \\
    \mathbf{H}^{(\ell)} &= \text{LayerNorm}(\mathbf{H}^{(\ell)} + \text{FFN}(\mathbf{H}^{(\ell)})),
    \end{align}
    where $\mathbf{W}_{Q_v}^{(\ell)}, \mathbf{W}_{K_v}^{(\ell)}, \mathbf{W}_{V_v}^{(\ell)} \in \mathbb{R}^{D \times D}$ are projection weight matrices, and all intermediate variables $\mathbf{Q}_v^{(\ell)}, \mathbf{K}_v^{(\ell)}, \mathbf{V}_v^{(\ell)}, \mathbf{A}_v^{(\ell)}, \mathbf{H}^{(\ell)} \in \mathbb{R}^{C \times D}$. Note that unlike the encoder's attention (which operates on temporal tokens), the decoder's attention operates on \textbf{variate tokens}, allowing each variable to attend to all other variables. The decoder output is projected and broadcast back to the original shape:
    \begin{equation}
    \mathbf{H}_{\text{out}} = \mathbf{H}^{(L_{\text{dec}})} \mathbf{W}_{\text{proj}} \in \mathbb{R}^{C \times D},
    \end{equation}
    \begin{equation}
    \tilde{\mathbf{Z}} = \mathbf{H}_{\text{out}} \otimes \mathbf{1}_P^\top \in \mathbb{R}^{C \times P \times D},
    \end{equation}
    where $\mathbf{W}_{\text{proj}} \in \mathbb{R}^{D \times D}$ is a projection weight matrix, $\otimes$ denotes replication broadcast along the patch dimension, and $\mathbf{1}_P \in \mathbb{R}^P$ is an all-ones vector.
    
    \textbf{Adaptive Residual Fusion.} To balance temporal and variate information, we introduce an adaptive residual connection:
    \begin{equation}
    \mathbf{Z}_{\text{fused}} = \mathbf{Z}_{\text{enc}} + \lambda \cdot \tilde{\mathbf{Z}},
    \end{equation}
    where $\lambda \in [0, 1]$ is a learnable or predefined residual weight. This enables the model to adaptively combine temporal patterns (e.g., trends, seasonality) from the encoder and variate dependencies (e.g., cross-variate correlations) from the inverted decoder. In our experiments, we observe that for \textbf{low-dimensional} datasets ($C \leq 10$), a smaller $\lambda$ (e.g., 0.3) prevents over-emphasizing limited variable interactions, while for \textbf{high-dimensional} datasets ($C \geq 100$), a larger $\lambda$ (e.g., 1.0) fully exploits rich cross-variate dependencies.
    
    \textbf{Prediction Head.} Finally, the fused representation is reshaped and linearly projected to generate the final predictions. For each variable $c \in \{1, \dots, C\}$, we flatten its $P$ patches of $D$-dimensional features into a single vector:
    \begin{equation}
    \mathbf{z}_{\text{flat},c} = \text{vec}(\mathbf{Z}_{\text{fused},c,:,:}) \in \mathbb{R}^{D \cdot P},
    \end{equation}
    where $\text{vec}(\cdot)$ denotes the vectorization operation that flattens a matrix column-wise. We stack the flattened vectors across all variables:
    \begin{equation}
    \mathbf{Z}_{\text{flat}} = [\mathbf{z}_{\text{flat},1}; \mathbf{z}_{\text{flat},2}; \dots; \mathbf{z}_{\text{flat},C}] \in \mathbb{R}^{C \times (D \cdot P)}.
    \end{equation}
    Finally, a linear layer projects to the prediction horizon:
    \begin{equation}
    \hat{\mathbf{Y}} = \mathbf{Z}_{\text{flat}} \mathbf{W}_{\text{head}} + \mathbf{1}_C \mathbf{b}_{\text{head}}^\top \in \mathbb{R}^{C \times H},
    \end{equation}
    where $\mathbf{W}_{\text{head}} \in \mathbb{R}^{(D \cdot P) \times H}$ is the projection weight matrix, $\mathbf{b}_{\text{head}} \in \mathbb{R}^H$ is the bias vector, and $\mathbf{1}_C \in \mathbb{R}^C$ is an all-ones vector for broadcasting the bias. Equivalently, the prediction for variable $c$ is:
    \begin{equation}
    \hat{\mathbf{y}}_c = \mathbf{W}_{\text{head}}^\top \mathbf{z}_{\text{flat},c} + \mathbf{b}_{\text{head}} \in \mathbb{R}^H,
    \end{equation}
    allowing each variable to independently map its temporal features to future predictions.
        
    \subsection{Comparison with Related Architectures}
    \label{subsec:comparison}
    
    InvDec differs from existing methods primarily in its hybrid design philosophy. Compared to \textbf{PatchTST}~\cite{nie2023timeseriesworth64}, InvDec explicitly models cross-variate dependencies by adding an inverted decoder, breaking the limitation of the channel-independence assumption. Unlike \textbf{iTransformer}~\cite{liu2024itransformerinvertedtransformerseffective}, which applies attention solely on the variate dimension in a pure inverted design, and \textbf{Crossformer}~\cite{zhang2023crossformer}, which interleaves temporal and variate modeling throughout the network via two-dimensional segmented embeddings, InvDec retains temporal encoding in the encoder, thereby preserving the ability to learn critical temporal patterns.
    
    Other cross-dimension modeling methods also explore correlations among variables, each with distinct emphases. \textbf{VCformer}~\cite{yang2024vcformervariablecorrelationtransformer} focuses on capturing lagged correlations among variables, explicitly modeling time delays to handle asynchronous dependencies. \textbf{UniTST}~\cite{liu2024unitsteffectivelymodelinginterseries} attempts to balance intra-series and inter-series dependencies, though its design is relatively complex. Furthermore, methods designed for spatial grid data, such as \textbf{GridTST}~\cite{cheng2024leveraging2dinformationlongterm}, leverage two-dimensional spatial information to model proximity relationships among grid-structured variables. While such methods excel in scenarios with explicit spatial topology (e.g., traffic networks, weather forecasting), they struggle to generalize to arbitrary multivariate time series where variables lack explicit spatial structure. In contrast, InvDec dynamically learns variable correlations through variate-level attention mechanisms without requiring predefined graph structures or spatial topology, offering broader applicability.
    
    It is also worth noting that \textbf{TimeXer}~\cite{wang2024timexer} similarly introduces learnable variable embeddings to handle heterogeneity among variables. However, TimeXer and InvDec differ fundamentally in their \textbf{application scenarios and methodologies}, making direct comparison difficult. First, TimeXer focuses on \textbf{exogenous forecasting tasks}, where the core objective is to leverage external covariates (e.g., weather, holidays, prices) to predict target variables, whereas InvDec targets \textbf{standard multivariate forecasting tasks}, predicting future values for all endogenous variables. This difference in task definition leads to fundamentally different data characteristics: TimeXer handles exogenous variables that often exhibit systematic time lags and heterogeneous physical measurement units, requiring variable-specific normalization and embedding at the input stage. In contrast, InvDec handles endogenous variables that, while correlated, emphasize \textbf{dynamic dependencies} rather than static heterogeneity. Second, in terms of \textbf{architectural design}, TimeXer adopts an end-to-end Transformer and fuses variable embeddings at the input layer (operating on the full $C \times P \times D$ sequence), whereas InvDec delays the introduction of variable embeddings until the decoder stage (operating only on the pooled $C \times D$ tensor), achieving a clear separation between temporal and variate modeling. TimeXer excels in domain-specific exogenous variable scenarios, while InvDec is better suited for capturing purer and more general endogenous dynamics. Therefore, the two methods differ substantially in task type, data assumptions, and application domains, and cannot be judged superior or inferior through direct performance comparison. Rather, they should be viewed as complementary approaches targeting different forecasting paradigms.
    
    InvDec's design philosophy maintains a \textbf{clear separation of concerns}: the encoder focuses on capturing local patterns in the temporal dimension (via the patching mechanism), while the decoder focuses on modeling global dependencies in the variate dimension (via variate-level attention). Furthermore, since variable embeddings are used only when necessary, InvDec is more efficient in high-dimensional scenarios, and its plug-and-play nature theoretically allows it to be combined with any temporal encoder.

    \section{Experiments}
    \label{sec:experiments}
    
    \subsection{Experimental Setup}
    \label{subsec:setup}
    
    We conduct extensive experiments to evaluate the effectiveness of InvDec across diverse multivariate time series forecasting benchmarks. Our evaluation focuses on three key questions: (1) Does InvDec outperform state-of-the-art baselines, particularly in high-dimensional scenarios? (2) How do individual components of InvDec contribute to its performance? (3) What insights can be gained from analyzing the learned representations and attention patterns? This section describes our experimental setup, including datasets, evaluation metrics, baseline methods, and implementation details.
    
    \textbf{Datasets.} We evaluate InvDec on seven widely-used benchmark datasets for long-term time series forecasting, all adopted from Autoformer~\cite{wu2022autoformerdecompositiontransformersautocorrelation} and other related models: \textbf{ETTh1, ETTh2, ETTm1, ETTm2} (Electricity Transformer Temperature datasets)~\cite{zhou2021informerefficienttransformerlong}, \textbf{Weather} (meteorological data), \textbf{Electricity} (electricity consumption), and \textbf{Traffic} (road occupancy rates). These datasets span diverse domains and exhibit varying characteristics in terms of dimensionality ($C$), sequence length, and temporal dynamics. Table~\ref{tab:datasets} summarizes the statistics of each dataset. Notably, Electricity ($C=321$) and Traffic ($C=862$) represent high-dimensional scenarios where cross-variate dependencies are particularly important, while ETT and Weather datasets have lower dimensions ($C \leq 21$). Each dataset is split into training (70\%), validation (10\%), and test (20\%) sets following standard practice. For preprocessing, we apply z-score normalization to each variable independently.
    
    \textbf{Evaluation Metrics.} Following standard practice in time series forecasting~\cite{nie2023timeseriesworth64,liu2024itransformerinvertedtransformerseffective}, we evaluate model performance using two metrics: Mean Squared Error (MSE) and Mean Absolute Error (MAE). MSE measures the average squared difference between predictions and ground truth, penalizing larger errors more heavily, while MAE provides a linear scale of average prediction error. Lower values indicate better performance for both metrics. We report the average results across multiple prediction horizons ($H \in \{96, 192, 336, 720\}$ for most datasets) to assess model robustness across different forecasting lengths.
    
    \textbf{Baselines.} We compare InvDec against nine state-of-the-art baselines spanning diverse architectural families: (1) \textbf{Transformer-based methods}: PatchTST~\cite{nie2023timeseriesworth64}, iTransformer~\cite{liu2024itransformerinvertedtransformerseffective}, Crossformer~\cite{zhang2023crossformer}, Autoformer~\cite{wu2022autoformerdecompositiontransformersautocorrelation}, and FEDformer~\cite{zhou2022fedformerfrequencyenhanceddecomposed}; (2) \textbf{MLP-based methods}: DLinear~\cite{zeng2022transformerseffectivetimeseries} and TimeMixer~\cite{wang2024timemixerdecomposablemultiscalemixing}; (3) \textbf{CNN-based methods}: TimesNet~\cite{wu2023timesnettemporal2dvariationmodeling} and ModernTCN~\cite{donghao2024moderntcn}. These baselines represent the current state-of-the-art across different modeling paradigms and provide comprehensive coverage of channel-independent, cross-dimension, and hybrid approaches. For most baselines, we directly report results from their original papers when available. For models requiring reimplementation or when results are unavailable, we reproduce them using the same experimental protocol and hyperparameters as InvDec to ensure fair comparison.
    
    \textbf{Implementation Details.} We implement InvDec using PyTorch and conduct all experiments on NVIDIA GPUs. Following standard practice~\cite{nie2023timeseriesworth64,wu2023timesnettemporal2dvariationmodeling}, we use a fixed lookback window of $L=96$ for all datasets and evaluate on four prediction horizons: $H \in \{96, 192, 336, 720\}$. For each dataset, we conduct four experiments corresponding to these horizons. The embedding dimension $D$, number of attention heads $N_h$, and number of encoder layers $L_{\text{enc}}$ vary across datasets and horizons to balance model capacity and computational efficiency. For InvDec-specific components, the inverted decoder uses $L_{\text{dec}}=2$ layers with $4$ attention heads across most settings. The residual weight $\lambda$ in adaptive residual fusion is set to 0.3 for low-dimensional datasets ($C \leq 21$) and 1.0 for high-dimensional datasets ($C \geq 100$). We train all models using the Adam optimizer with a learning rate of $10^{-3}$ and batch sizes ranging from 32 to 128 depending on the dataset and prediction horizon. To ensure fair comparison, all baselines reproduced by us use identical shared hyperparameters (e.g., $L$, $H$, batch size) as InvDec for the same dataset. Due to the complexity of hyperparameter configurations across different datasets and horizons, we refer readers to our code repository for complete experimental scripts and settings.
    
    \begin{table}[t]
    \centering
    \caption{Statistics of the seven benchmark datasets.}
    \label{tab:datasets}
    \begin{tabular}{lccc}
    \toprule
    \textbf{Dataset} & \textbf{Variables ($C$)} & \textbf{Frequency} & \textbf{Domain} \\
    \midrule
    ETTh1 & 7 & Hourly & Energy \\
    ETTh2 & 7 & Hourly & Energy \\
    ETTm1 & 7 & 15-min & Energy \\
    ETTm2 & 7 & 15-min & Energy \\
    Weather & 21 & 10-min & Meteorology \\
    Electricity & 321 & Hourly & Energy \\
    Traffic & 862 & Hourly & Transportation \\
    \bottomrule
    \end{tabular}
    \end{table}
        
    \subsection{Main Results}
    \label{subsec:main_results}
    
    Table~\ref{tab:main_results} presents the average performance of InvDec-PatchTST compared to nine state-of-the-art baselines across seven benchmark datasets. The results are averaged over four prediction horizons ($H \in \{96, 192, 336, 720\}$) for each dataset.
    
    \begin{table*}[t]
    \centering
    \caption{Average forecasting performance (MSE and MAE) across four prediction horizons ($H \in \{96, 192, 336, 720\}$). For ETT, results are averaged over all four datasets (ETTh1, ETTh2, ETTm1, ETTm2). \textbf{Best}, \underline{second-best}, and \dashuline{third-best} results are highlighted.}
    \label{tab:main_results}
    \resizebox{\textwidth}{!}{
    \begin{tabular}{lcccccccc}
    \toprule
    \multirow{2}{*}{\textbf{Model}} & \multicolumn{2}{c}{\textbf{Weather}} & \multicolumn{2}{c}{\textbf{Electricity}} & \multicolumn{2}{c}{\textbf{Traffic}} & \multicolumn{2}{c}{\textbf{ETT}} \\
    \cmidrule(lr){2-3} \cmidrule(lr){4-5} \cmidrule(lr){6-7} \cmidrule(lr){8-9}
    & MSE & MAE & MSE & MAE & MSE & MAE & MSE & MAE \\
    \midrule
    \textbf{InvDec-PatchTST} & \textbf{0.247} & \textbf{0.275} & \textbf{0.173} & \underline{0.276} & \underline{0.468} & \underline{0.303} & \dashuline{0.384} & \dashuline{0.402} \\
    PatchTST & \dashuline{0.258} & \underline{0.280} & 0.219 & 0.298 & \dashuline{0.481} & \dashuline{0.304} & \textbf{0.381} & \textbf{0.397} \\
    iTransformer & \underline{0.260} & \dashuline{0.281} & \underline{0.178} & \textbf{0.270} & \textbf{0.428} & \textbf{0.282} & \underline{0.383} & \underline{0.399} \\
    Crossformer & 0.263 & 0.322 & 0.244 & 0.334 & 0.550 & \dashuline{0.304} & 0.685 & 0.598 \\
    Autoformer & 0.337 & 0.389 & 0.227 & 0.338 & 0.628 & 0.379 & 0.531 & 0.490 \\
    FEDformer & 0.309 & 0.360 & 0.214 & 0.327 & 0.610 & 0.376 & 0.432 & 0.453 \\
    Non-Stationary & 0.288 & 0.314 & 0.193 & 0.296 & 0.624 & 0.340 & 0.521 & 0.484 \\
    TimesNet & \dashuline{0.258} & 0.285 & \dashuline{0.192} & \dashuline{0.295} & 0.620 & 0.336 & 0.416 & 0.404 \\
    DLinear & 0.265 & 0.317 & 0.212 & 0.300 & 0.625 & 0.383 & 0.442 & 0.419 \\
    \bottomrule
    \end{tabular}
    }
    \end{table*}
    
    \textbf{Overall Performance.} InvDec-PatchTST demonstrates significant advantages on high-dimensional datasets where cross-variate dependencies are critical. On Electricity ($C=321$), InvDec achieves the best MSE with 20.9\% and 2.8\% improvements over PatchTST and iTransformer respectively. On Traffic ($C=862$), InvDec ranks second with 2.7\% improvement over PatchTST. On Weather ($C=21$), InvDec achieves the best performance on both metrics with 4.3\% MSE improvement over PatchTST. These results validate that the inverted decoder effectively exploits cross-variate relationships when sufficient variables are present~\cite{qiu2025comprehensivesurveydeeplearning}. On low-dimensional ETT ($C=7$), InvDec achieves third-best performance with minimal gaps (0.8\% MSE behind PatchTST), demonstrating that the adaptive residual fusion successfully balances temporal and variate modeling without prohibitive overhead when cross-channel dependencies are limited.

    \textbf{Comparison with Recent Methods.} Compared to iTransformer~\cite{liu2024itransformerinvertedtransformerseffective}, which applies self-attention exclusively on the variate dimension, InvDec achieves superior performance on medium-dimensional datasets (2.8\% on Electricity, 5.0\% on Weather) while iTransformer excels on ultra-high-dimensional Traffic where $C \gg L$. On low-dimensional ETT, both achieve nearly identical performance (<0.3\% difference). This validates InvDec's design: by retaining temporal encoding in the encoder while applying variate attention in the decoder, InvDec achieves favorable balance across diverse dimensionalities~\cite{liu2024itransformerinvertedtransformerseffective}. Crossformer~\cite{zhang2023crossformer}, which interleaves temporal and variate modeling, consistently underperforms InvDec, suggesting clear separation of concerns is superior to entangled processing.
    
    \textbf{Comparison with Traditional Architectures.} MLP-based DLinear~\cite{zeng2022transformerseffectivetimeseries} struggles on high-dimensional Electricity (22.5\% worse than InvDec), confirming simple linear mappings cannot capture complex cross-variate dependencies. CNN-based TimesNet~\cite{wu2023timesnettemporal2dvariationmodeling} achieves competitive results but remains inferior to InvDec on high-dimensional datasets, suggesting limitations in modeling long-range cross-variate interactions through convolutions. Traditional Transformer methods (Autoformer~\cite{wu2022autoformerdecompositiontransformersautocorrelation}, FEDformer~\cite{zhou2022fedformerfrequencyenhanceddecomposed}, Non-Stationary~\cite{liu2023nonstationarytransformersexploringstationarity}) generally lag behind recent innovations~\cite{kim2025comprehensivesurveydeeplearning}, indicating decomposition-based approaches do not adequately address cross-variate dependencies in high-dimensional scenarios.
        
    \subsection{Ablation Study}
    \label{subsec:ablation}
    
    We conduct ablation studies on the Weather dataset to validate the effectiveness of each component in InvDec-PatchTST. Table~\ref{tab:ablation} presents the results across four prediction horizons.
    
    \begin{table}[t]
    \centering
    \caption{Ablation study on Weather dataset. We report MSE across four prediction horizons ($H \in \{96, 192, 336, 720\}$). Lower is better.}
    \label{tab:ablation}
    \begin{tabular}{lccccc}
    \toprule
    \textbf{Configuration} & \textbf{96} & \textbf{192} & \textbf{336} & \textbf{720} & \textbf{Avg} \\
    \midrule
    \multicolumn{6}{l}{\textit{InvDec Necessity}} \\
    \quad w/o InvDec & 0.177 & 0.225 & 0.278 & 0.354 & 0.259 \\
    \quad w/ InvDec & \textbf{0.160} & \textbf{0.211} & \textbf{0.272} & \textbf{0.347} & \textbf{0.247} \\
    \midrule
    \multicolumn{6}{l}{\textit{Number of Layers}} \\
    \quad 1-layer & 0.163 & 0.212 & 0.269 & 0.345 & 0.248 \\
    \quad 2-layer & \textbf{0.160} & \textbf{0.211} & \textbf{0.272} & \textbf{0.347} & \textbf{0.247} \\
    \midrule
    \multicolumn{6}{l}{\textit{Number of Heads}} \\
    \quad 1-head & 0.162 & 0.209 & 0.267 & 0.347 & 0.246 \\
    \quad 2-head & 0.162 & 0.212 & 0.269 & 0.348 & 0.248 \\
    \quad 4-head & \textbf{0.160} & \textbf{0.211} & \textbf{0.272} & \textbf{0.347} & \textbf{0.247} \\
    \midrule
    \multicolumn{6}{l}{\textit{Residual Weight $\lambda$}} \\
    \quad $\lambda$=0.0 & 0.177 & 0.225 & 0.278 & 0.354 & 0.259 \\
    \quad $\lambda$=0.5 & 0.164 & 0.213 & 0.272 & 0.351 & 0.250 \\
    \quad $\lambda$=1.0 & \textbf{0.160} & \textbf{0.211} & \textbf{0.272} & \textbf{0.347} & \textbf{0.247} \\
    \bottomrule
    \end{tabular}
    \end{table}
    
    \textbf{Necessity of InvDec.} Setting $\lambda=0$ disables InvDec completely, making the model equivalent to PatchTST baseline. Comparing w/o InvDec ($\lambda=0$) with the full model ($\lambda=1$), we observe consistent improvements across all prediction horizons with an average MSE reduction of 4.6\% (from 0.259 to 0.247). This validates that the inverted decoder effectively captures cross-variate dependencies even on the moderate-dimensional Weather dataset ($C=21$). The gains are particularly notable at shorter horizons (9.6\% at $H=96$, 6.2\% at $H=192$), suggesting that cross-variate modeling provides immediate benefits for near-term forecasting.
    
    \textbf{Impact of Layer Depth.} Reducing InvDec from 2 layers (the default configuration) to 1 layer results in nearly identical average performance (0.247 vs 0.248), with minimal differences across individual horizons (<0.5\%). This indicates that for moderate-dimensional datasets such as Weather, a single decoder layer is sufficient to capture cross-variate dependencies. However, we retain 2 layers as the default configuration to provide additional modeling capacity for high-dimensional datasets where more complex variable interactions may require deeper representations.
    
    \textbf{Impact of Attention Heads.} Using 1 head achieves comparable performance to 4 heads (the default configuration) , i.e., 0.246 vs 0.247 average MSE, demonstrating that even a single attention head of InvDec can effectively capture cross-variate patterns on Weather. With 2 heads, performance slightly decreases to 0.248. This counter-intuitive result suggests that for moderate-dimensional scenarios, excessive attention heads may introduce redundancy or optimization challenges. However, we maintain 4 heads as the default to ensure robust performance across diverse datasets, particularly high-dimensional ones where multiple attention heads can capture distinct correlation patterns among different variable subsets.
    
    \textbf{Impact of Residual Weight $\lambda$.} Setting $\lambda=0.5$ (weaker InvDec contribution) results in 1.2\% performance degradation compared to $\lambda=1.0$ (from 0.247 to 0.250 average MSE). This demonstrates that balanced fusion of temporal encoding (from encoder) and variate modeling (from InvDec) is optimal for Weather. The $\lambda=0.5$ configuration shows consistent degradation across all horizons, confirming that the inverted decoder's full contribution ($\lambda=1.0$) is necessary to maximize cross-variate modeling benefits. This validates our design choice of using $\lambda=1.0$ for medium-to-high dimensional datasets where cross-variate dependencies are substantial.
        
    \subsection{Model Analysis}
    \label{subsec:model_analysis}
    
    We conduct comprehensive analysis to understand InvDec-PatchTST's computational characteristics, scalability across different dimensionalities, and performance trends across prediction horizons.
    
    \subsubsection{Computational Complexity Analysis}
    
    The computational complexity of InvDec-PatchTST with representative baselines is compared in Table~\ref{tab:complexity}. Following standard practice~\cite{nie2023timeseriesworth64,liu2024itransformerinvertedtransformerseffective}, we analyze complexity in terms of the number of time steps $L$, number of variables $C$, model dimension $d$, and number of patches $P = L/S$ where $S$ is the stride.
    
    \begin{table}[h]
    \centering
    \caption{Complexity comparison of InvDec with related methods. $C$: number of variables, $L$: sequence length, $P$: number of patches ($P \ll L$), $d$: embedding dimension.}
    \label{tab:complexity}
    \begin{tabular}{lcc}
    \toprule
    \textbf{Method} & \textbf{Time Complexity} & \textbf{Space Complexity} \\
    \midrule
    PatchTST & $\mathcal{O}(C \cdot P^2 \cdot d)$ & $\mathcal{O}(C \cdot P \cdot d)$ \\
    iTransformer & $\mathcal{O}(C^2 \cdot L \cdot d)$ & $\mathcal{O}(C^2 + C \cdot L \cdot d)$ \\
    Crossformer & $\mathcal{O}(C \cdot P^2 \cdot d + C^2 \cdot P \cdot d)$ & $\mathcal{O}(C \cdot P \cdot d + C^2 \cdot P)$ \\
    \textbf{InvDec (Ours)} & $\mathcal{O}(C \cdot P^2 \cdot d + C^2 \cdot d)$ & $\mathcal{O}(C \cdot P \cdot d + C^2)$ \\
    \bottomrule
    \end{tabular}
    \end{table}
    
    From a computational complexity perspective, InvDec's encoder has time complexity $\mathcal{O}(C \cdot P^2 \cdot d)$ for temporal self-attention across $C$ variables, while the inverted decoder adds $\mathcal{O}(C^2 \cdot d)$ for variate self-attention. The overall time complexity is therefore $\mathcal{O}(C \cdot P^2 \cdot d + C^2 \cdot d)$. Compared to iTransformer's $\mathcal{O}(C^2 \cdot L \cdot d)$, InvDec is significantly more efficient thanks to patch-based compression ($P \ll L$) and global pooling that reduces temporal complexity. For typical configurations where $P \gg C$ (e.g., Weather with $P \approx 12, C=21$), the $\mathcal{O}(C^2 \cdot d)$ overhead is modest. The space complexity $\mathcal{O}(C \cdot P \cdot d + C^2)$ is also favorable, with the decoder attention matrix being the main additional memory cost. Empirically, InvDec adds only 2-5\% overhead to training time while achieving significant performance gains on high-dimensional datasets.
    
    \subsubsection{Performance across Dimensionalities}
    
\begin{figure}[h]
    \centering
    \includegraphics[width=0.9\columnwidth]{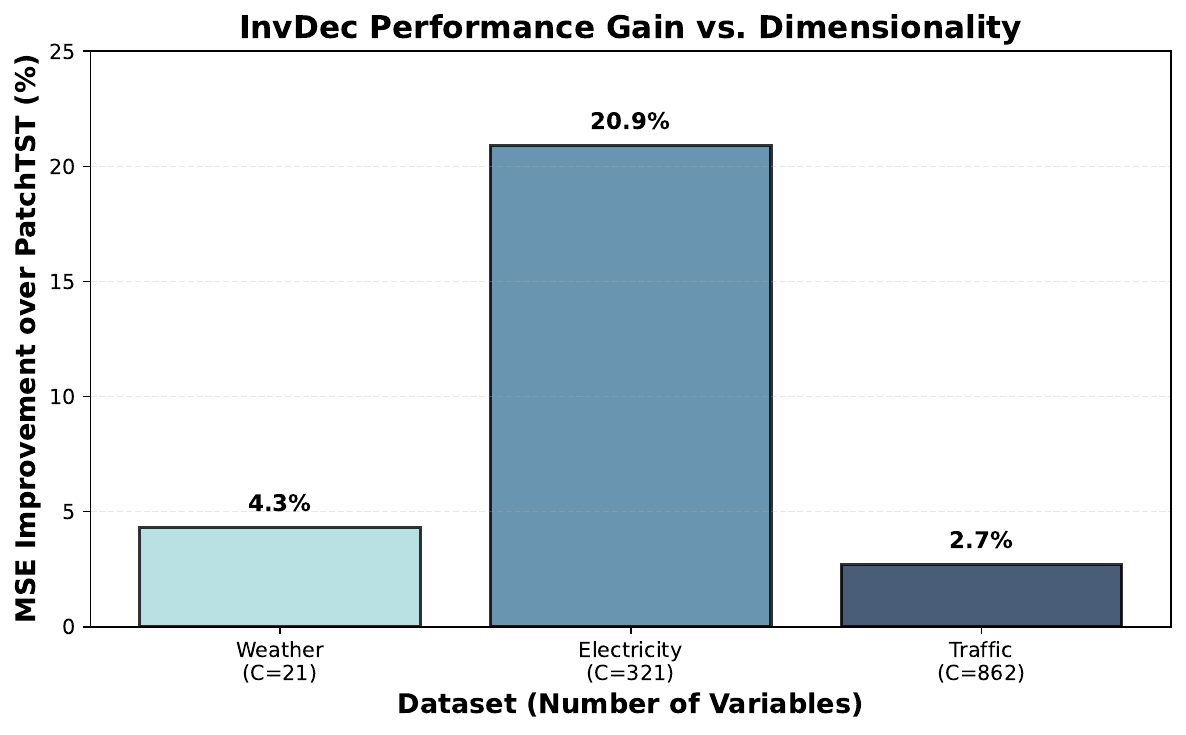}
    \caption{Relative MSE improvement (\%) of InvDec-PatchTST over PatchTST baseline across datasets with different dimensionalities. InvDec's advantage increases with the number of variables, validating its design for high-dimensional scenarios.}
    \label{fig:dimensionality}
    \end{figure}

    It is clearly illustrated in Figure~\ref{fig:dimensionality} how InvDec's performance advantage varies with dataset dimensionality. We plot the relative MSE improvement over PatchTST baseline across 3 datasets with different, large numbers of variables.
    
    The results reveal a clear trend: InvDec's improvement over channel-independent baseline grows with dimensionality, validating our core hypothesis that cross-variate modeling becomes increasingly important as the number of variables increases~\cite{qiu2025comprehensivesurveydeeplearning}. On Weather ($C=21$), the improvement increases to 4.3\%, demonstrating that even moderate dimensionality benefits from variate attention. On Electricity ($C=321$), InvDec achieves substantial 20.9\% improvement, confirming that the inverted decoder effectively exploits rich cross-variate dependencies in high-dimensional settings. On Traffic ($C=862$), the gain stabilizes at 2.7\%, suggesting a trade-off between modeling capacity and ultra-high dimensionality where iTransformer's pure variate-attention architecture may be more suitable~\cite{liu2024itransformerinvertedtransformerseffective}. This dimensionality-performance relationship validates InvDec's design philosophy: combining temporal encoding with adaptive variate modeling provides optimal balance across diverse scenarios.
    
    \subsubsection{Impact of Prediction Horizon}
    
    The performance of InvDec across different prediction horizons on the Electricity dataset is analyzed in Table~\ref{tab:horizon_analysis}, which exhibits strong cross-variate dependencies due to its high dimensionality ($C=321$).
    
    \begin{table}[h]
    \centering
    \caption{MSE performance on Electricity dataset across prediction horizons. InvDec maintains consistent advantages over baselines at all horizons, with particularly strong gains at longer horizons.}
    \label{tab:horizon_analysis}
    \begin{tabular}{lcccc}
    \toprule
    \textbf{Model} & \textbf{96} & \textbf{192} & \textbf{336} & \textbf{720} \\
    \midrule
    PatchTST & 0.201 & 0.201 & 0.215 & 0.257 \\
    iTransformer & 0.148 & 0.162 & 0.178 & 0.225 \\
    \textbf{InvDec-PatchTST} & \textbf{0.146} & \textbf{0.164} & \textbf{0.180} & \textbf{0.203} \\
    \midrule
    \textit{Improvement vs. PatchTST} & 27.4\% & 18.4\% & 16.3\% & 21.0\% \\
    \textit{Improvement vs. iTransformer} & 1.4\% & -1.2\% & -1.1\% & 9.8\% \\
    \bottomrule
    \end{tabular}
    \end{table}
    
    InvDec demonstrates consistent improvements across all prediction horizons, with average 20.8\% gain over PatchTST. The improvement pattern reveals interesting insights: at short horizon (96 steps), InvDec achieves 27.4\% improvement over PatchTST, showcasing the immediate benefit of cross-variate modeling. At medium horizons (192-336 steps), the gain stabilizes around 16-18\%, while at long horizon (720 steps), it increases to 21.0\%. This U-shaped trend suggests that cross-variate dependencies are valuable both for capturing immediate correlations and for long-term trend modeling. Compared to iTransformer, InvDec achieves comparable performance at short-medium horizons but significantly outperforms at the longest horizon (9.8\% improvement at 720 steps), demonstrating that the hybrid temporal-variate architecture is particularly advantageous for long-term forecasting where both temporal patterns and cross-variate relationships are critical.
    
    \section{Conclusion}
    \label{sec:conclusion}
    
    In this paper, we introduced InvDec (Inverted Decoder), a novel hybrid architecture for multivariate time series forecasting that achieves principled separation between temporal encoding and variate-level decoding. By combining a patch-based temporal encoder with an inverted decoder that operates on the variate dimension, InvDec addresses a fundamental limitation of existing methods: the inability to simultaneously capture fine-grained temporal patterns and complex cross-variate dependencies without sacrificing one for the other.
    
    Our key contributions are threefold. First, we proposed the InvDec architecture, which to the best of our knowledge represents the first work to combine a temporal encoder with a variate-level decoder in this manner, enabling clear separation of modeling concerns while maintaining computational efficiency. Second, we introduced a delayed variate embedding strategy that enriches variable-specific representations without interfering with temporal encoding, alongside an adaptive residual fusion mechanism ($\lambda$) that dynamically balances temporal and variate information across datasets of varying dimensions. Third, we conducted extensive experiments on seven widely-used benchmarks, demonstrating that InvDec-PatchTST achieves significant performance gains on high-dimensional datasets—including a 20.9\% MSE reduction on Electricity (321 variables) and 4.3\% improvement on Weather compared to PatchTST baseline—while maintaining competitive performance on low-dimensional datasets such as ETT.
    
    Our experimental findings reveal several important insights. First, the advantage of cross-variate modeling grows significantly with dataset dimensionality, validating the hypothesis that explicit modeling of variable interactions becomes increasingly critical as the number of variables increases~\cite{qiu2025comprehensivesurveydeeplearning}. Second, the hybrid temporal-variate architecture demonstrates superior long-term forecasting capability compared to pure channel-independent or pure variate-attention methods, particularly at prediction horizons of 720 steps where InvDec achieves 21.0\% improvement over PatchTST on Electricity. Third, ablation studies confirm that each component of InvDec—the inverted decoder, multi-head variate attention, and adaptive residual fusion—contributes meaningfully to overall performance, with the residual weight $\lambda$ serving as an effective mechanism for adapting to datasets of different dimensions.
    
    \textbf{Limitations.} Despite these contributions, InvDec has several limitations that warrant discussion. First, on ultra-high-dimensional datasets such as Traffic ($C=862$), where the number of variables far exceeds the number of patches ($C \gg P$), pure variate-attention architectures such as iTransformer~\cite{liu2024itransformerinvertedtransformerseffective} still demonstrate advantages. This suggests that when cross-variate dependencies completely dominate temporal patterns, dedicated variate-centric designs may be more suitable. Second, the inverted decoder introduces additional computational overhead of $\mathcal{O}(C^2 \cdot d)$, which can become non-trivial for extremely high-dimensional scenarios (e.g., $C > 1000$). While this overhead is modest for datasets considered in this work, scalability to even larger variable spaces remains an open challenge. Third, due to computational resource constraints, we were unable to fully explore the impact of model scale (e.g., embedding dimension, number of layers) on InvDec's performance. It is possible that larger model capacities could further improve results, particularly on complex high-dimensional datasets.
    
    \textbf{Future Work.} Several promising directions emerge from this work. First, exploring more efficient variate modeling mechanisms—such as sparse attention patterns, hierarchical variable clustering, or learnable graph structures—could reduce the $\mathcal{O}(C^2)$ complexity while retaining modeling capacity. Second, developing adaptive architecture selection strategies that automatically choose between channel-independent, hybrid, and pure variate-attention designs based on dataset characteristics (e.g., dimensionality, correlation structure) could yield more robust forecasting systems. Third, extending InvDec to other time series analysis tasks such as classification, anomaly detection, and imputation could demonstrate the generality of the temporal-variate separation principle. Finally, investigating the integration of InvDec with domain-specific knowledge (e.g., spatial topology in traffic networks, hierarchical relationships in economic indicators) represents an exciting avenue for further improving forecasting accuracy in specialized applications.
    
    In conclusion, InvDec demonstrates that principled separation of temporal and variate modeling, combined with adaptive fusion mechanisms, provides an effective framework for multivariate time series forecasting across diverse scenarios. We hope this work inspires further research into hybrid architectures that leverage the complementary strengths of different modeling perspectives.

    \section*{Code and Data Availability}

    The code for InvDec-PatchTST and all experimental scripts will be made publicly soon. All benchmark datasets used in this work (ETT, Weather, Electricity, Traffic) are publicly available and details on data sources are provided in Section~\ref{subsec:setup}.
    
    \bibliographystyle{ieeetr}
    \bibliography{Reference}
\end{document}